\documentclass[letterpaper, 10 pt, conference]{ieeeconf}  

\IEEEoverridecommandlockouts                              

\usepackage{comment} 
\usepackage{graphicx} 
\usepackage{bm} 
\usepackage{amsmath} 
\usepackage{amsfonts} 
\let\labelindent\relax 
\usepackage{bbm} 
\usepackage{siunitx} 
\usepackage{enumitem}
\usepackage{booktabs}
\usepackage{multirow}
\usepackage{hyperref}
\usepackage{cite}

\usepackage{caption} 
\newcommand{\vect}[1]{\mathbf{#1}}              
\newcommand{\nR}[1]{\mathbb{R}^{#1}}        
\newcommand{\nN}[1]{\mathbb{N}^{#1}}		
\newcommand{\matr}[1]{\mathbf{#1}}		        
\newcommand{\mtime}{t}

\newcommand{\robot}{\text{robot}}
\newcommand{\configRobot}{\vect{q}}        
\newcommand{\dconfigRobot}{\dot{\vect{q}}} 
\newcommand{\configObject}{\vect{o}}       
\newcommand{\dconfigObject}{\dot{\vect{o}}}
\newcommand{\point}{\vect{p}}
\newcommand{\orientation}{\matr{R}}
\newcommand{\pose}{\matr{T}}

\newcommand{\reward}{J}
\newcommand{\objTarget}{\configObject^*}
\newcommand{\cinput}{\vect{u}}
\newcommand{\aff}{A}
\newcommand{\act}{\alpha}
\newcommand{\orprop}{Q}
\newcommand{\mppiMode}{m}
\newcommand{\pfeature}{\vect{f_p}}
\newcommand{\pcd}{\mathcal{C}}

\newcommand{\state}{\vect{x}}
\newcommand{\stateobs}{\vect{\hat{x}}}

\newcommand{\controller}{\kappa}
\newcommand{\noise}{\vect{z}}

\newcommand{\jointtorque}{\vect{\tau}}
\newcommand{\ddconfigRobot}{\ddot{\vect{q}}}
\newcommand{\massMatrix}{\matr{M}(\configRobot)}
\newcommand{\coriolis}{\matr{Co}(\configRobot, \dconfigRobot)}

\newcommand{\error}[1]{\tilde{#1}}
\newcommand{\desired}[1]{#1^*}
\newcommand{\configRobotError}{\error{\vect{q}}} 
\newcommand{\dconfigRobotError}{\dot{\error{\vect{q}}}} 
\newcommand{\configRobotDesired}{\desired{\configRobot}} 
\newcommand{\dconfigRobotDesired}{\desired{\dconfigRobot}} 
\newcommand{\ddconfigRobotDesired}{\desired{\ddconfigRobot}} 
\newcommand{\citet}{\cite} 
\newcommand{\citep}{\cite} 

\title{\LARGE \bf
Learning Agent-Aware Affordances for Closed-Loop Interaction with Articulated Objects.
}

\author{Giulio Schiavi$^{*}$, Paula Wulkop$^{*}$, Giuseppe Rizzi, Lionel Ott, Roland Siegwart, Jen Jen Chung$^{\dag}$
 \thanks{This project has received funding from the European Union’s Horizon 2020 research and innovation programme under grant agreement No 101017008 (Harmony).}%
 \thanks{$^{*}$ Equal contribution}%
 \thanks{Authors are with the Autonomous Systems Lab,
        ETH Zurich, Switzerland. Email: pwulkop@ethz.ch, giulio.schiavi@outlook.com.}%
\thanks{ $^\dag$Also with the School of ITEE, The University of Queensland, Australia.}
\thanks{Project page: \url{https://paulawulkop.github.io/agent_aware_affordances}}
 }

\begin{document}

\maketitle
\thispagestyle{empty}
\pagestyle{empty}

\begin{abstract}

    Interactions with articulated objects are a challenging but important task for mobile robots. To tackle this challenge, we propose a novel closed-loop control pipeline, which integrates manipulation priors from affordance estimation with sampling-based whole-body control. We introduce the concept of agent-aware affordances which fully reflect the agent's capabilities and embodiment and we show that they outperform their state-of-the-art counterparts which are only conditioned on the end-effector geometry. Additionally, closed-loop affordance inference is found to allow the agent to divide a task into multiple non-continuous motions and recover from failure and unexpected states. Finally, the pipeline is able to perform long-horizon mobile manipulation tasks, i.e. opening and closing an oven, in the real world with high success rates (opening: $71\%$, closing: $72\%$).

\end{abstract}

\section{INTRODUCTION}

\label{sec:introduction}

In the future, autonomous mobile robots could relieve humans of tedious, repetitive manual tasks in a wide variety of environments like hospitals, homes or laboratories. Many daily tasks require interaction with articulated objects, for example to open the door of a dishwasher. This is especially challenging for a robotic agent, because of the complex system dynamics caused by the object's degrees of freedom and kinematic constraints.

A common approach is to estimate the kinematic and semantic properties of articulated objects from visual data \cite{learning2019abbatematteo, articulated-object-pose-estimation, screwnet,shape2motion2019wang} and leverage this information for planning and control \cite{articulated2021mittal,arduengo}. This two-staged approach often requires heuristics, for example defining the grasping point on the handle, limiting the flexibility to deal with unseen articulation types and object geometries. A more generic approach using affordances, i.e. where and how an agent can interact with an object, was recently explored \citet{where2act2021mo,vatmart2021wu}. Given a point cloud of an articulated object, they use neural networks to predict point-wise interaction scores (\emph{actionability}), which are used as priors for a downstream robotic controller. While these approaches show promising initial results, they neglect that affordances are always dependent on the agent's capabilities, which are defined by the hardware and the controller. Instead, the state-of-the-art models are trained on a disembodied gripper, disregarding the robot kinematics and joint limits. This can lead to the predictions of motions that are infeasible for the real robot. Furthermore, these pipelines query the affordance module only once and then keep the interaction pose and planned trajectory fixed. Due to this open-loop perception and planning setup, they cannot perform long-term tasks requiring a change of interaction pose when the robot reaches kinematic or joint limits.

\begin{figure} [t]
   \centering
   \includegraphics[width=0.35\textwidth]{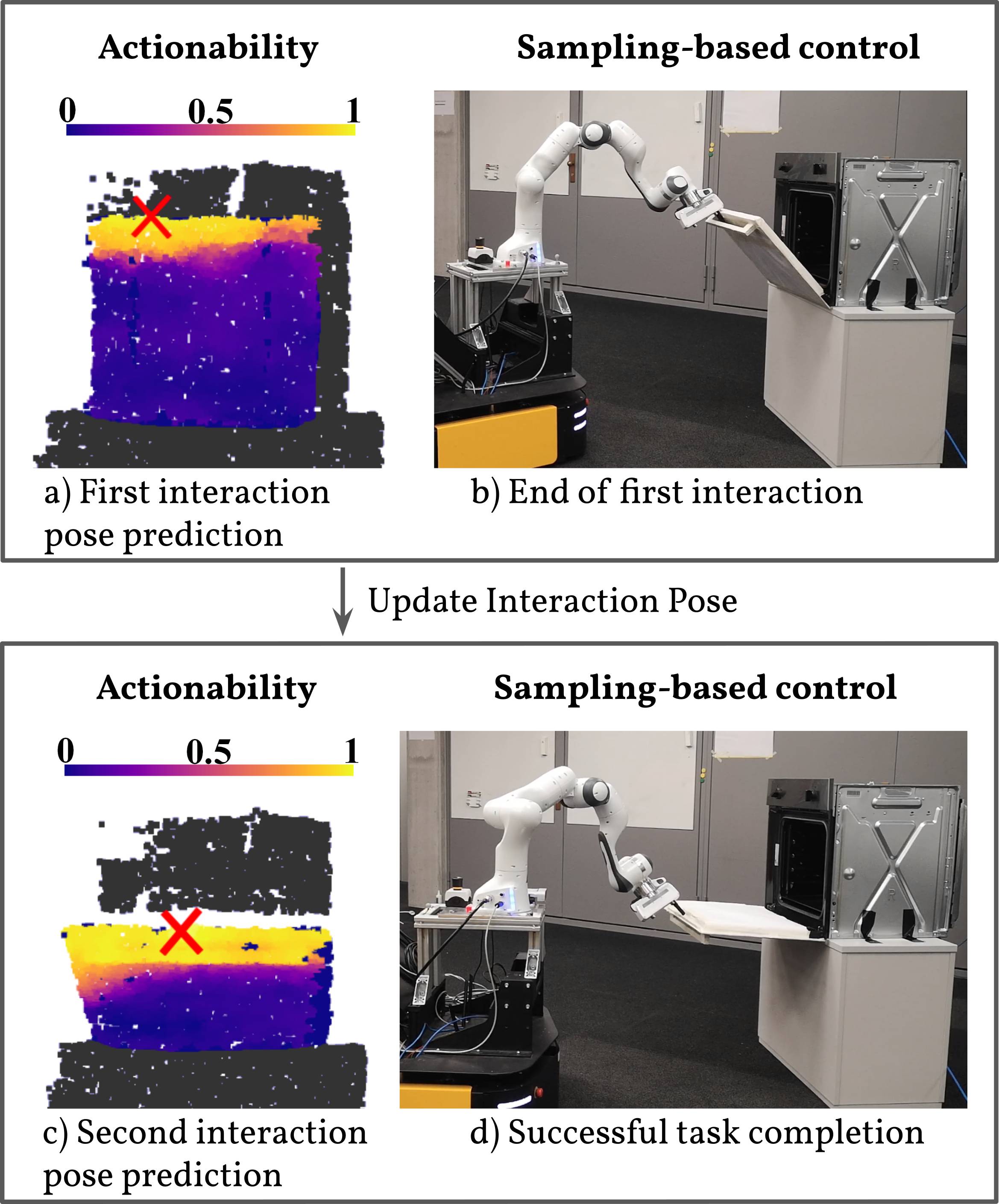}
   \caption{Real-world experiment of opening an oven in two motions. a) \& c): Estimated actionability map where the red cross represents the selected interaction point. b) The first interaction pose becomes unfavorable, therefore an update is triggered. d): Successful task completion after the second interaction.}
   \label{image:realworld_interaction}
 \vspace*{-3mm}

\end{figure}

To overcome these limitations, we propose a novel closed-loop pipeline combining agent-aware affordance perception with sampling-based whole-body control. Concretely, this work deals with the non-prehensile manipulation of articulated objects with one degree of freedom using a mobile manipulator equipped with a single, fixed cylindrical finger. Taking inspiration from the \emph{Where2Act} framework \cite{where2act2021mo}, we train an artificial neural network to estimate point-level affordances from visual data, indicating the success likelihood for interactions at each point. The pose proposal with the highest affordance score is passed on to a sampling-based controller (based on \cite{information_theoretic_mpc,giuseppe_mppi}), which then iteratively determines the best interaction trajectory. This setup is well suited for non-prehensile manipulation since the interaction location and trajectory are continuously adapted based on real-time feedback, enabling adaptive and robust task execution. In contrast to previous end-effector-aware approaches, we train our affordance inference network with data collected using the full model of our target robot platform, making our predictions fully agent-aware. While this implies the need to repeat data collection and training for a different agent, we find that it significantly improves the number of successful interactions. Additionally, unlike previous approaches with a fixed interaction pose, our pipeline is able to re-evaluate the affordance model at any point during task execution, which allows the robot to execute long-term tasks where a change of interaction pose is required. 
In this work, we show in ablation experiments that agent-aware training significantly increases the quality of pose proposals, and that allowing the agent to change the interaction pose during the task increases the robustness of the integrated affordance-control pipeline. Additionally, we benchmark our method against VAT-Mart~\cite{vatmart2021wu} as a state-of-the-art work and we perform experiments in the real world~(Fig.~\ref{image:realworld_interaction}), reaching a success rate of more than 70\% for both fully opening and closing an oven door using a mobile manipulator. In summary, our contributions are:
\begin{itemize}[noitemsep,topsep=0pt]
    \item We formulate the concept of agent-aware object affordances which are conditioned on the full shape and kinematics of the robot. 
    \item We propose a novel closed-loop manipulation framework that combines the concept of visual affordances with a sampling-based controller.
\end{itemize}

\section{RELATED WORK}
\label{sec:related_work}

\paragraph*{Control strategies for manipulating articulated objects}
A common approach for the manipulation of articulated objects is to extract object properties like part segmentation and joint kinematics from visual data \cite{articulated-object-pose-estimation,object-size-estimation,learning2019abbatematteo, screwnet, shape2motion2019wang, kineverse2022} and use this to plan an interaction trajectory \cite{sturm2010, martinmartin2019, jain2020}. Mittal et al.~\citet{articulated2021mittal} recently successfully used this approach by implementing a Model Predictive Controller (MPC) to track the feed-forward trajectory plan. However, MPC struggles with discontinuities caused by contact dynamics, such that a simplification of the problem is required, e.g. splitting the task into first achieving a stable grasp at a fixed interaction position and then executing a predefined motion based on known articulation kinematics. To alleviate these limitations, sampling-based control has recently emerged, allowing a more task-specific and complex interaction representation \cite{mppi_control, information_theoretic_mpc, stein_var_mpc}. In previous work by Rizzi et al.~\citet{giuseppe_mppi}, a sampling-based controller was successfully applied as a whole-body closed-loop controller for the non-prehensile manipulation of articulated objects. However, to decrease the size of the sampling space, a fixed end-effector interaction pose still has to be provided a priori for each object. 

\paragraph*{Affordances for manipulating articulated objects}
An affordance is defined as the ability of an agent to perform an action with a target object in a given environment~\cite{ecological1979gibson,affordances2020ardn}. In robotics, the concept of affordances is commonly applied either on an object-level, e.g. a dishwasher is \emph{openable}~\cite{learning2020nagarajan, li2022ifrexplore}, or on a point-level to encode information about the object geometry and grasp possibilities~\cite{learning2018zeng,pohl2020affordance,learning2020yenchen, o2oafford2021}. In the current literature, affordances are usually learned from labeled visual data \cite{interaction_tensor, 3daffordancenet, Myers2015AffordanceDO} or from self-supervised interactions \cite{ada_afford, o2oafford2021, learning2020nagarajan, li2022ifrexplore}. Recent works have applied this concept to the manipulation of articulated objects. VAT-Mart~\citet{vatmart2021wu} and Where2Act~\citet{where2act2021mo} propose frameworks which learn affordances for robotic manipulation from interactions generated in a photo-realistic simulator \cite{sapien2020xiang}. A network is then trained to infer interaction points and trajectory proposals for downstream manipulation tasks from visual data. However, these approaches generate training data by simulating the interaction between a disembodied end-effector and a unit-sphere scaled object. This results in affordances that are not only unaware of the kinematics and control of the actual robot platform, but also of the realistic object size. Consequently, the trajectories derived from these approaches might be infeasible in real life, e.g. due to joint limits or collisions of the arm with the object. Additionally, they only query the visual model at the beginning of the interaction in an open-loop fashion, keeping the planned grasp location and trajectory constant during the interaction. UMPNet~\citet{umpnet_xu_2022} on the other hand proposed a closed-loop affordance-based manipulation framework that can update the interaction direction during task execution. However, it still uses a fixed interaction point, determined at the start of the interaction using only end-effector-aware (agent-agnostic) affordances. 

\section{PROBLEM FORMULATION}
\label{sec:problem_formulation}

We consider a physical system consisting of a mobile manipulator and an articulated object which is composed of two rigid bodies connected by a rotational or translational joint. We define the state $\state = [\configRobot, \configObject, \dconfigRobot, \dconfigObject]$ as the configurations of the robot and object and their time derivatives, where $\configRobot$ consists of the position of the mobile base relative to the object as well as the joint angles of the manipulator. The robot kinematics and the object geometry and articulation are assumed to be provided a priori. Given a desired object configuration $\objTarget$, the point cloud from the robot's point of view $\pcd$ and the current state observation $\state$, the overall objective is to find a robot control policy that generates joint velocity commands $\cinput$ such that $\configObject = \objTarget$. 



\begin{figure*}[t]
    \centering
    \includegraphics[width=.76\textwidth]{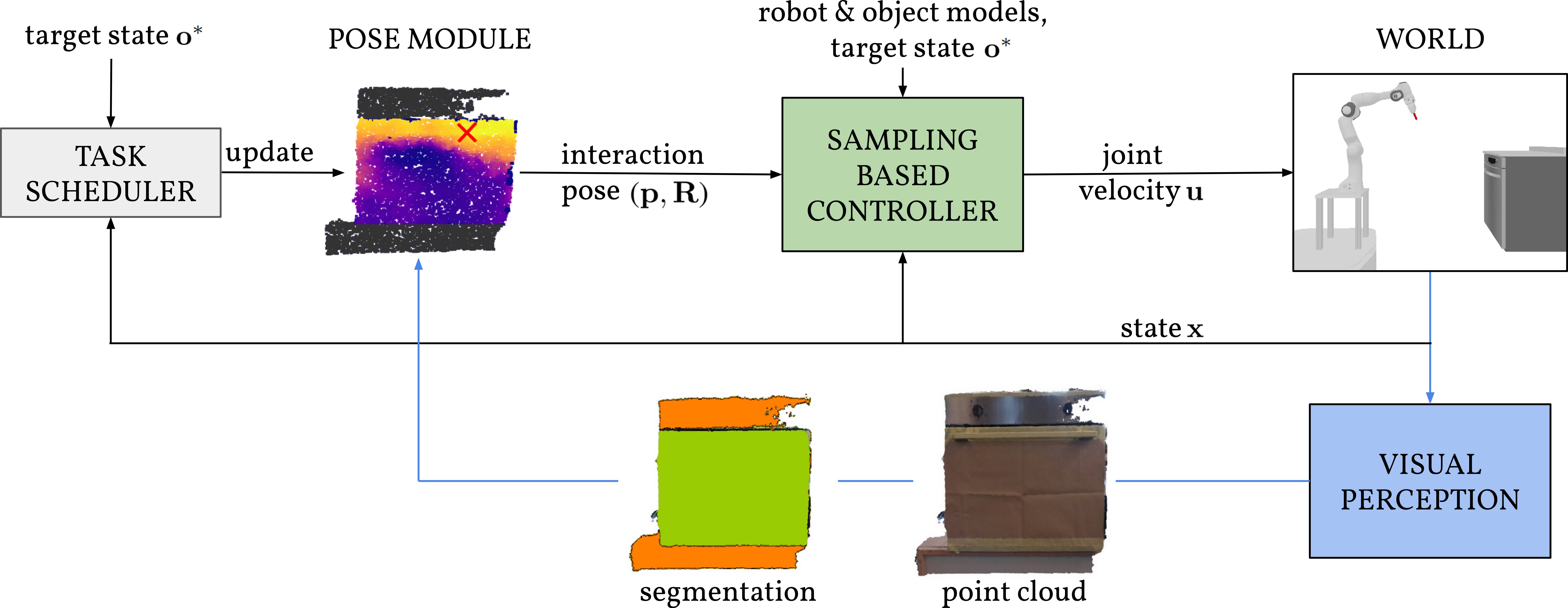}
    \caption{Block diagram of the control pipeline. The pose module uses affordance estimation to choose the optimal end-effector reference pose. The sampling-based controller generates velocity commands to interact with the object. The task scheduler can trigger an update of the interaction pose if required.}
    \label{img:pipeline_block_diagram}
\end{figure*}

\section{AGENT-AWARE AFFORDANCE LEARNING}
\label{sec:method}
We propose a novel control pipeline combining an agent-aware affordance network with a sampling-based whole-body controller (Fig.~\ref{img:pipeline_block_diagram}). Given an object point cloud $\pcd$, from which the movable object part is extracted by applying a segmentation mask, and target configuration $\objTarget$, we use an affordance neural network to select a favorable end-effector interaction pose (point $\point \in \mathbb{R}^3$ and orientation $\orientation \in SO(3)$). The interaction pose is then used by the sampling-based controller $\controller$ to generate joint velocity commands $\cinput = \controller(\state \mid \objTarget, \point, \orientation)$. When the interaction fails or stops leading to any improvement of $\configObject$, a task scheduler triggers a pose update as described in Section~\ref{subsubsec:control_pipeline}.

\subsection{Affordance-based Pose Module}
\label{subsubsec:pose_module}
\paragraph*{Pose inference}
\label{subsubsec:affordance_definition}
 Let $\reward(\state_{0:N}, \controller(\state_{0:N-1}) \mid \objTarget)$ be a reward function that, given the target object configuration $\objTarget$, maps a trajectory of system states $\{\state_{0:N}\}$ and controller inputs $\{\cinput_{0:N-1}\}$ to a binary reward value. Given an object and its target configuration $\objTarget$, we define the affordance $\aff$ of an interaction pose ($\point, \orientation$) as the expected value of $\reward$ over all the state-input trajectories induced by the controller $\controller(\state \mid \objTarget, \point, \orientation)$ from initial state $\state_0$. Inferring the optimal interaction pose by maximizing the affordance function $\aff$ is impractical, as the set of all possible interaction positions and orientations is extremely large. We therefore follow \cite{where2act2021mo,vatmart2021wu} and define two auxiliary functions. The orientation proposal function $\orprop$ generates end-effector orientation proposals given an interaction point. The actionability function $\act$ models the point-wise expected affordance $\aff$ over a distribution of interaction orientations. At test time, we first select the interaction point that maximizes $\act$. We then generate multiple orientation proposals and score their affordance. Finally, the highest-scoring pose proposal ($\point, \orientation$) is selected and passed on to the sampling-based controller. 

\begin{figure}[tb]
\centering
\includegraphics[width=0.49\textwidth]{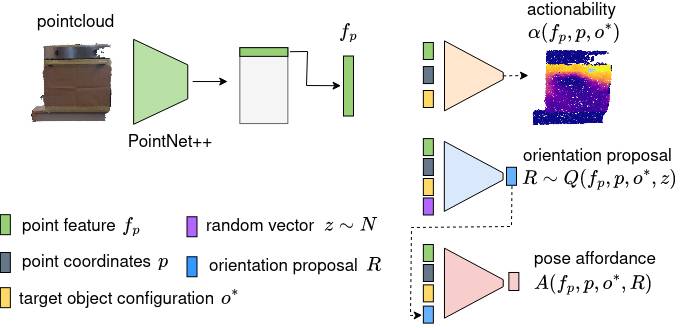}
\caption{The pose module, based on \cite{where2act2021mo}, combines a PointNet++ \cite{qi2017pointnetplusplus} feature encoder with three MLP decoding heads, outputting actionability $\act$, orientation proposal $\orprop$ and pose affordance $A$.}
\label{img:network_architecture}
\end{figure}

\paragraph*{Network architecture}
\label{subsubsec:network_architecture}
We use a neural network based on \cite{where2act2021mo} to learn the functions required by the optimization problem, namely affordance $\aff$, orientation proposal $\orprop$ and actionability $\act$. The network architecture, shown in Fig.~\ref{img:network_architecture}, combines a PointNet++~\cite{qi2017pointnetplusplus} feature encoder with three Multilayer Perceptron (MLP) decoding heads, one for each function. All three networks take as input the point position $\point$, its feature vector $\pfeature$ and the target object configuration $\objTarget$. The orientation proposal module additionally takes a sampled Gaussian noise vector $\noise \in \nR{10} \sim \mathcal{N}(\vect{0}, \matr{1})$ as input to generate proposals $\orientation$ which are then passed on as an input to the affordance prediction module.

\paragraph*{Data collection}
\label{subsubsec:data_collection}
The training data is collected in a simulation setup where we simulate the complete control pipeline as well as the kinematic chain, collision bodies and dynamics of the target robot platform in order to create agent-aware affordances. A baseline agent samples a random interaction pose $(\point, \orientation)$ on the movable part of the object, where the orientations are sampled only within a $45^\circ$ cone of the surface normal at point $\point$ (based on \cite{vatmart2021wu}). Given the sampled pose and the desired configuration $\objTarget$, the sampling-based controller attempts to interact with the object for 10s. A binary reward $\reward$ rates the interaction as successful if the object configuration was changed by more than $5^\circ$ towards the target $\objTarget$. Finally, we balance the dataset to contain the same number of successful and unsuccessful samples. One advantage of using the sampling-based controller to generate training data is the relatively high rate of successful interactions (13.6\%), which makes it much more data efficient than first having to train a control policy (as done in \cite{vatmart2021wu}). 

\paragraph*{Training and losses}
\label{subsubsec:training}
Given the point cloud and the interaction sample $(\point, \orientation, \objTarget, \reward)$, we train the components of the pose module jointly where the affordance module learns to predict the reward realization $\reward$ using a binary cross-entropy loss, while the orientation proposal module uses the mean cosine similarity loss on quaternion predictions. The actionability module is trained on an L1 loss, where the ground truth actionability for a given pose proposal can be calculated using predictions from the affordance and orientation modules. We use an Adam optimizer and employ learning rate scheduling and early stopping to ensure convergence. Training a network takes around $3$ hours on a low-grade GPU (NVIDIA GTX $1050$ Ti with $4$ GB RAM).

\subsection{Control Pipeline}
\label{subsubsec:control_pipeline}

\paragraph*{Sampling-based controller}
The sampling-based controller uses a physics engine to forward simulate the system dynamics of the robot and object to iteratively refine the motion trajectory~\citep{giuseppe_mppi}. At every time step, it samples multiple robot joint velocity references $\cinput$ around the current best guess and simulates the resulting system dynamics for a time horizon of 1s. The best trajectory is then selected based on a cost function, which is made up of multiple components. The object cost encodes the offset from the target $\objTarget$, the collision cost punishes robot-object collision, the joint limit cost and the arm reach cost prevent mechanically unreachable configurations. Finally, the pose reach cost is defined as the distance between the end-effector and the reference pose $(\point, \orientation)$, which is required to make the optimization problem feasible. A low-level proportional-integral controller then converts the velocity reference $\cinput$ to the executed joint torques commands.

\paragraph*{Task scheduler}
To increase the robustness of the system, we implement a closed-loop affordance inference setup, allowing the robot to change its interaction pose when the current one is no longer favorable. In practice, the task scheduler triggers a pose update whenever the object state cost stagnates. To update the pose, the robot moves to a configuration from where it can observe the full object, records a point cloud, segments it and passes it through the pose module to obtain the new interaction pose.

%
%

\section{EXPERIMENTS}
\label{sec:result}
To evaluate the performance of our pipeline, we conducted three different experiments in simulation and additionally validated our approach in real life. For training, we collect for each task $12000$ simulations with eleven different object models from the PartNet-Mobility dataset categories \textit{oven} and \textit{dishwasher}, which both have a revolute joint~\cite{partnet2019mo}. First, we compared affordances conditioned only on the end-effector (as in \cite{where2act2021mo, vatmart2021wu}) against affordances conditioned on the full robot. Second, we evaluated the advantage of allowing the agent to update the interaction pose when the current one is no longer advantageous, i.e. split one task into multiple non-continuous motions. These two experiments were performed with seven unseen object models from the training categories since they are especially challenging in terms of reach and collision and are therefore well suited to study the advantages of our method. Finally, we show that our approach is also capable of generalizing to unseen object categories and articulation types. To this end, we tested our pipeline on 140 object models from the categories \emph{safe}, \emph{table} and \emph{washing machine} containing both revolute and prismatic joints and compare it to VAT-Mart~\cite{vatmart2021wu} as a state-of-the-art benchmark.

\subsection{Agent and Environment}
The robot platform consists of a seven degree-of-freedom manipulator mounted on a mobile base. The manipulator hand is equipped with a single, fixed cylindrical finger. One of the advantages of this simple finger design is that it allows the physics engine RAISIM to simulate the contact dynamics, which is a computationally demanding task~\cite{contactdynamics2020}, smoothly in real-time with a simulation timestep of $0.0015\,$s. To generate the pointclouds, the robot and the object are rendered in the SAPIEN environment \cite{sapien2020xiang}, where the object's initial position and scale are sampled uniformly at random within predefined, category-specific bounds.

\subsection{Baselines}
As a minimum baseline, we employ the same random agent used during training for data collection. To evaluate the importance of agent-aware affordances, we compare against a purely end-effector-aware baseline where we collected the training data with a disembodied, freely moving gripper initialized directly at the interaction pose and thus ignoring the issue of reach. This is similar to the affordance-learning paradigm in Where2Act~\citep{where2act2021mo}. As an additional baseline, we enhance the end-effector-aware network with a reachability filter using a Jacobian-based inverse kinematics (IK) check as well as with a robot-object collision filter. These filters are implemented such that each interaction pose proposed by the end-effector-aware network has to pass the feasibility check before being executed. Next, to evaluate the effect of enabling a change of interaction pose, we created for both the end-effector-aware and the agent-aware network one version where the pose stays fixed during the interaction and a closed-loop setup that allows pose updates.

\subsection{Results}
\paragraph*{Agent-aware vs end-effector-aware}
We first evaluate the quality of the proposed agent-aware interaction poses with a simplified version of the pipeline which keeps the interaction pose fixed. As in \cite{where2act2021mo}, we report the sample success rate as the percentage of interaction poses creating movement towards the target $\objTarget$, i.e. $|\Delta \configObject|\geq 5^\circ$, as well as the sample reach rate defined as the percentage of pose references that are reachable by the agent. The considered tasks are to \textit{open} or \textit{close} articulated objects from the testing dataset using the full robot model. Over the 500 trials of each task, the initial object configuration $\configObject_0$ is such that half of the trials start with the object fully closed or fully open (for the \emph{open} and \emph{close} tasks, respectively), the other half start with the object open to an angle sampled randomly over the full joint range.

\begin{figure} [tb]
  
\centering
\includegraphics[width=0.43\textwidth]{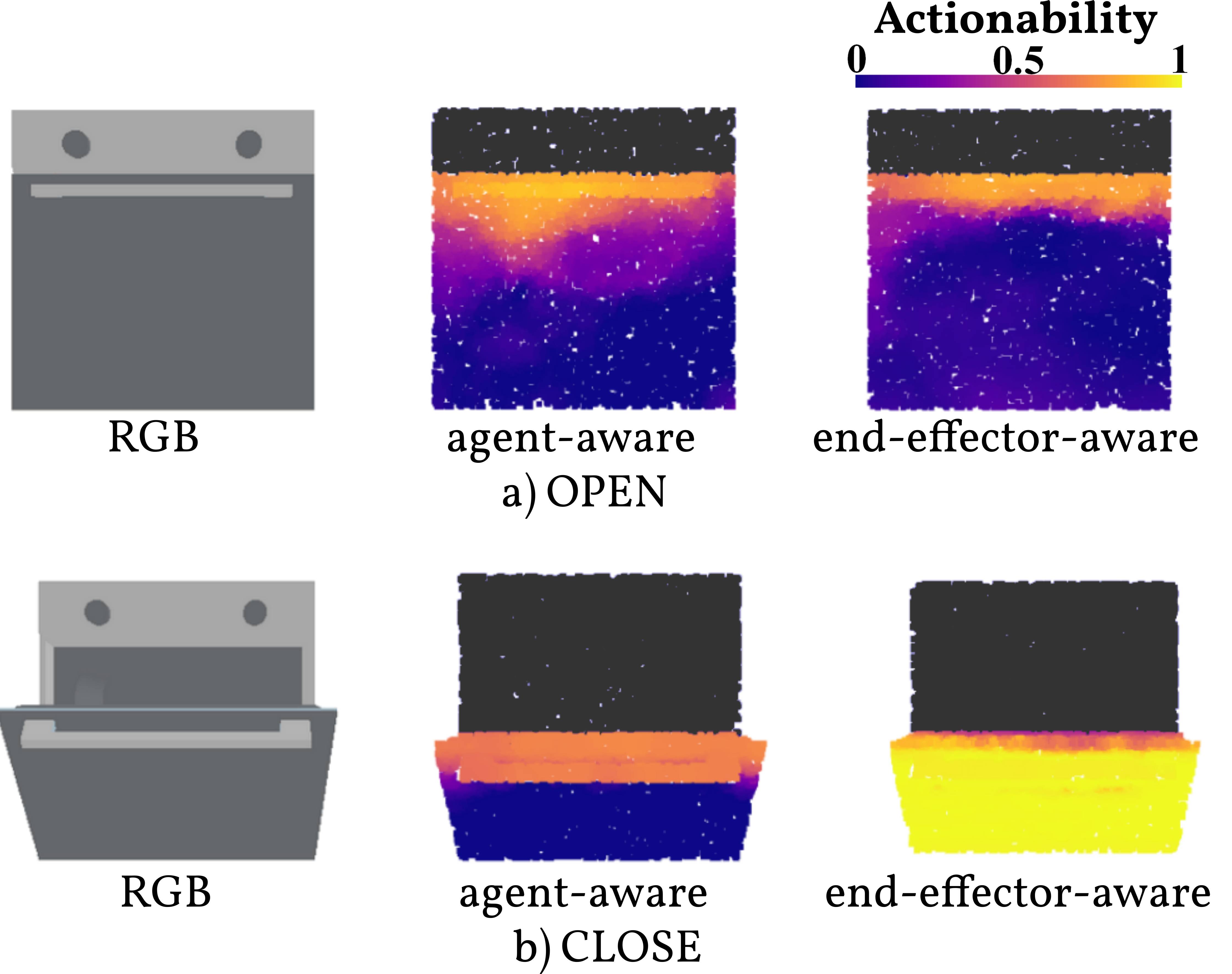}
 \caption{a) \emph{Open} task: both networks learn to interact with a similar set of points. b) \emph{Close} task: the agent-aware network strongly prefers reachable points while the end-effector-aware network, which was trained on a disembodied gripper, also predicts points on the door surface which are unreachable for the full agent.}
\label{image:close_open_hand_v_full}

 \end{figure}

In the \emph{open} task, both networks learned to interact with a similar set of points (Fig.~\ref{image:close_open_hand_v_full}a), leading to similar average performance (Table~\ref{table:hand_v_full_open}). In the \emph{close} task on the other hand, while the end-effector-aware network considers the entire outer surface of the door to be actionable, the agent-aware network strongly prefers interaction points close to the top edge (Fig.~\ref{image:close_open_hand_v_full}b). As the end-effector-aware network was trained on a disembodied gripper, it often predicts interaction poses that are unreachable for the full agent due to kinematic or collision constraints. This results in a significantly lower reach and success rate compared to the agent-aware version.

Table~\ref{table:hand_v_full_open} shows that even though the feasibility checks progressively improve the reachability and success rate of poses, the performance is still lower than for our proposed agent-aware network. From this experiment we conclude that there is no simple heuristic filter that can restore the agent-awareness a posteriori. The failure cases of the agent-aware model on the other hand are mostly due to suboptimal actionability predictions of the network, i.e. it proposes poses that can be reached but do not result in successful interactions. Additionally, we compared multiple network instances trained with different random seeds and found that while stochasticity has a visible effect on the affordance map, the resulting performance is consistent across networks. In general, we observe that the sampling-based controller provides a certain robustness against small errors of the predicted interaction pose, as it locally adapts the pose while minimizing the controller cost.

\begin{table}[t]
    \begin{minipage}{\linewidth}

      \centering
        \footnotesize
        \begin{tabular}{l l l}
        \toprule
      
        OPEN & s. success r. & s. reach r.\\
                              \midrule
         Random               & 15.5\% & 68.5\% \\
         EE-aware             & 91.9\% & 99.8\% \\
         \bf{Agent-aware} & \bf{92.1\%} & \bf{99.9\%} \\
         \bottomrule
               CLOSE & s. success r. & s. reach r.\\
        \midrule
         Random               & 21.0\% & 61.5\% \\
         EE-aware             & 38.5\% & 54.2\% \\
         EE-aware + IK check             & 51.7\% & 65.0\% \\
         EE-aware + IK + collision check & 59.5\% & 76.6\% \\

         \bf{Agent-aware} & \bf{76.1\%} & \bf{92.0\%} \\
         \bottomrule
        \end{tabular}

    \caption{Sample success rate and sample reach rate using the full robot model for interaction trials of the \emph{open} and \emph{close} task. We compare our agent-aware network to a random baseline, the end-effector-aware network (EE-aware) and enhanced versions of the EE-aware network combined with an inverse kinematic (IK) and a collision check.}
    \label{table:hand_v_full_open}
    \label{table:hand_v_full_close}
    \end{minipage}

\end{table}

\paragraph*{Closed-loop pose update}
In this experiment we evaluate the performance on the full opening ($\configObject_0 = 0^\circ, \objTarget = 90^\circ$) and closing ($\configObject_0 = 90^\circ, \objTarget = 0^\circ$) tasks and report the task success rate. Over 1000 trials we compare our closed-loop pipeline, which allows the robot to change its interaction pose during the task, with an ablated fixed pose version. The agent-aware closed-loop pipeline achieves an extremely good task success rate of 88.6\% on both \textit{open} and \textit{close} tasks (Table~\ref{table:closed_loop_eval}). We observe two advantages of the closed-loop setup: Firstly, when an initially successful interaction pose becomes infeasible at an intermediate state, the task scheduler triggers an interaction pose update. Secondly, when the sampling-based controller is unable to perform the task at all by interacting with the current reference pose, the agent is able to recover by trying again at a different reference pose. In this simulated setup, we predominantly observed the second effect, because the sampling-based controller is able to locally adapt the interaction pose such that the task can be fulfilled in one motion, e.g. by sliding or rotating. Finally, the closed-loop setup also improves the results of the end-effector-aware network, but it is still outperformed by the agent-aware network.

\begin{table}[tb]
 \setlength{\tabcolsep}{5pt}
   \begin{minipage}{\linewidth}
      \centering
      \footnotesize
        \begin{tabular}{l l l | l l}
        \toprule
               & \multicolumn{2}{c|}{OPEN}  & \multicolumn{2}{c}{CLOSE}\\      
               & Agent-aware  & EE-aware & Agent-aware  & EE-aware \\ 
                              \midrule
        Closed-loop            & \bf{88.6\%}   & 76.6\%   & \bf{88.6\% }   & 64.3\%  \\
         
         Fixed pose              & 83.6\%   & 64.7\%  & 69.2\%    & 44.1\%   \\
         
         \bottomrule
        \end{tabular}
        \caption{Task success rate of the agent-aware and the end-effector-aware pipeline for the long-horizon open and close tasks. While the closed-loop setup improves the performance for both pipelines, the agent-aware version clearly outperforms its ablation.}
        \label{table:closed_loop_eval}
    \end{minipage}%
\end{table}

\begin{figure*} [ht]
   \centering
   \includegraphics[width=0.89\textwidth]{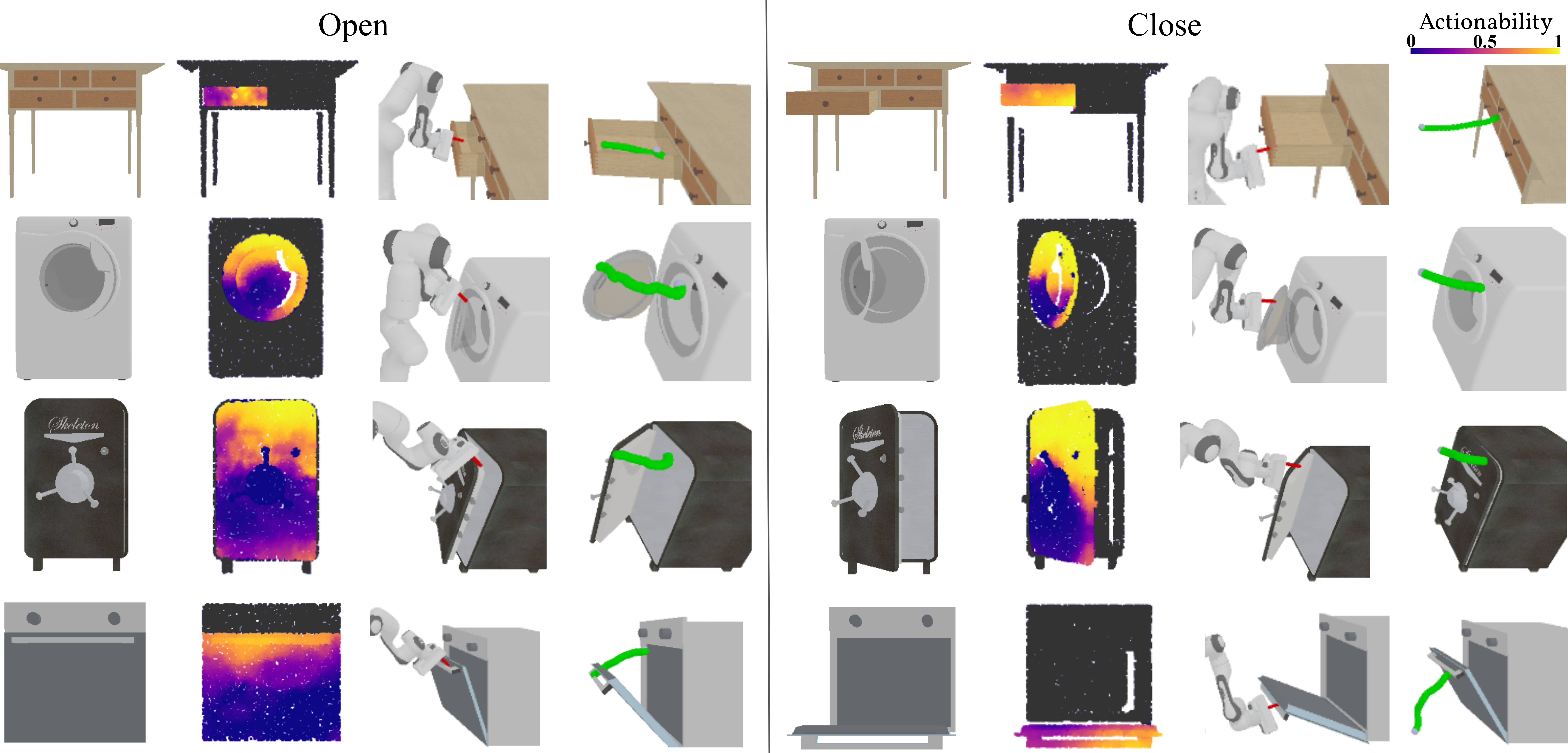}
   \caption{Qualitative analysis of the interactions with different articulated objects. Each interaction block shows from left to right: The RGB image, the predicted actionability map, the robot interacting with the object and the full trajectory of the end-effector (shown in green).}
   \label{image:vatmart_objects_results}

\end{figure*}

\paragraph*{Generalization capabilities}
To benchmark against VAT-Mart \cite{vatmart2021wu}, we evaluated our pipeline on their test objects (\emph{safe}, \emph{table} and \emph{washing machine}) and use their task definition which is to sample the target $\objTarget$ and the initial object configuration $\configObject_0$ randomly over the full joint space. Qualitative results in Fig.~\ref{image:vatmart_objects_results} show that our pipeline learned interaction strategies that generalize well to unseen object categories despite the limited object variability seen during training. Table~\ref{table:vat_mart_eval} shows that we outperform the VAT-Mart benchmark in all tasks and categories, even when only using a fixed interaction pose. Our pipeline is explicitly designed to profit from available information about the object and the robot which VAT-Mart does not use, namely the collision shape and articulation model as well as the real-time feedback of robot joints and articulation angle which are required by the controller. For our targeted service robotics applications, object CAD models are often readily available~\cite{IKEA}, while generating novel models would also only require a one-time effort.

\begin{table}[hb]
 \setlength{\tabcolsep}{2pt}
   \begin{minipage}{\linewidth}
      \centering
      \footnotesize 
        \begin{tabular}{l l l l l}
        \toprule
                                & close door &  open door & close drawer & open drawer \\ 
                              \midrule
         VAT-Mart~\citep{vatmart2021wu}                & 36.5\%    & 14.3\%     & 38.3\%       & 31.1\% \\
         
         Ours (Fixed Pose)        & 44.0\%    & 49.0\%     & 68.0\%       & 46.5\% \\
         
         \bf{Ours (Closed-loop)} & \bf{65.8\%} & \bf{66.7}\%     & \bf{80.5}\%       & \bf{59.1\%} \\ 
         \bottomrule
        \end{tabular}
        \caption{Task success rate, defined according to~\cite{vatmart2021wu} as within a tolerance of 15\% of the task, of our framework on unseen object categories compared to an ablated fixed pose version of our method and the VAT-Mart benchmark.}
        \label{table:vat_mart_eval}
    \end{minipage}%
\end{table}

\subsection{Real-world experiments}
We deployed our pipeline on a mobile manipulator interacting with an oven (Fig.~\ref{image:realworld_interaction}). Collision bodies and articulation type of the oven are provided to the controller a priori, while the articulation joint angle is obtained by integrating the angular velocity measurement of an IMU mounted on the back of the oven door. The pose of the robotic platform is tracked by an OptiTrack motion capture system and the point cloud data is collected from an onboard Azure Kinect sensor. The reflective glass surface of the oven door was covered to make it visible to the RGB-D camera. For part segmentation, the Iterative Closest Point (ICP) algorithm is used to match the real-world point cloud to the point cloud rendered in simulation, allowing a projection of the segmentation mask.

The closed-loop pipeline was tested over 25 trials of both the \textit{open} ($\configObject_0 = 0^\circ$, success if $\configObject_N \geq 75^\circ$) and \textit{close} ($\configObject_0 \geq 75^\circ$, success if $\configObject_N \leq 2^\circ$) tasks, where the joint limit of the real oven was at $85^\circ$. In Table~\ref{table:real_world_task_metrics}, real-world results are compared to simulation trials with the same setup and testing object. We report the task success rate (\textit{task s. r.}), the average number of interaction attempts (\textit{n. int.}) and the average number of successful motions ($|\Delta \configObject| \geq 5^\circ$) needed to complete a task (\textit{n. mot.}). 

\paragraph*{Results}
The network is able to produce reasonable actionability maps and interaction poses from real-world point cloud data (Fig.~\ref{image:realworld_interaction} and supplementary video). For the \textit{close} task, the behavior transfers very well from simulation to the real world: The robot is able to complete the task $72\%$ of the time and often in one motion (Table~\ref{table:real_world_task_metrics}). In the \textit{open} task, unlike in simulation, the real robot typically needs multiple attempts to slide the finger into the handle due to errors introduced by the state estimation and model mismatches. Once the end-effector is placed such that it can exert enough force, it opens the door to an intermediate state. This end-effector pose usually becomes unfavorable at a certain articulation angle (as seen in Fig.~\ref{image:realworld_interaction}b), such that a pose update is triggered and the robot uses a second and sometimes third motion to fully open the door. The recovery capabilities of the closed-loop pipeline are therefore essential to overcome the described sim-to-real gap, such that we achieve an overall task success rate of $71\%$ for opening.

\begin{table}[ht]
    \begin{minipage}{\linewidth}
      \centering
      \footnotesize
        \begin{tabular}{l l l l l}
        \toprule
         OPEN            & task s. r. & time (s) & n. int. & n. mot.\\ 
                              \midrule
         real        & 71\%  & 112$\pm$41 & 4.3$\pm$1.6 & 3.0$\pm$0.9\\
         sim         & 100\%  & 50$\pm$52 & 2.0$\pm$1.6  & 1.0$\pm$0.2 \\
         \midrule
         CLOSE                 & task s. r. & time (s) & n. int. & n. mot.\\ 
                              \midrule
         real        & 72\%  & 70$\pm$53 & 2.1$\pm$1.6 & 1.4$\pm$0.7\\
         sim         & 93\%  & 39$\pm$29 & 1.9$\pm$1.2  & 1.2$\pm$0.5 \\
         \bottomrule
        \end{tabular}
    \end{minipage}%
    \caption{Results from real-world and simulation testing for fully \textit{opening} and \textit{closing} an oven. The average number of interaction attempts (\textit{n. int.}) shows how often the task scheduler triggers a pose update and the number of motions (\textit{n. mot.}) is the subset of interactions that leads to an improvement of the articulation state.} 
    \label{table:real_world_task_metrics}
\end{table}


\section{CONCLUSIONS}
\label{sec:conclusion}

We introduced a novel closed-loop manipulation pipeline combining point-level affordance inference from visual data with whole-body control. The pipeline is based on sampling-based control, using affordances to estimate the end-effector interaction pose. Current state-of-the-art methods only condition affordance predictions on end-effector geometry. In contrast, we show that our agent-aware network is able to exploit the full robot model and low-level controller to improve the quality of the inferred interaction priors. Our pipeline also enables the agent to re-evaluate the pose model and update the interaction pose at any time during task execution. This allows the agent to split a task into two or more non-continuous motions and recover from failure and unexpected states. We find this to be crucial especially in the real-world experiments, partially compensating for the effects of the sim-to-real gap and allowing our pipeline to perform long-horizon mobile manipulation tasks with high success rates.

\paragraph*{Limitations and future work}
The sampling-based controller used in our pipeline requires precise measurements of the object joint state, currently provided by an external sensor (IMU). While this is not an issue in simulation and in controlled real-world environments, for general deployment the object pose estimation should be performed using onboard sensors, e.g. from RGB-D data \cite{asdf, articulated-object-pose-estimation, weng2021captra, pavlasek2020, desingh2019}. The controller also requires a precise object model, which could be estimated from visual data and optionally refined using interaction data \cite{screwnet, visualzeng2021, ditto}. Additionally, in future work the pipeline could be extended to allow for grasping, since not all tasks can be solved through non-prehensile manipulation.






\clearpage
\bibliographystyle{IEEEtran}
\bibliography{IEEEabrv, references}

\begin{thebibliography}{10}
\providecommand{\url}[1]{#1}
\csname url@rmstyle\endcsname
\providecommand{\newblock}{\relax}
\providecommand{\bibinfo}[2]{#2}
\providecommand\BIBentrySTDinterwordspacing{\spaceskip=0pt\relax}
\providecommand\BIBentryALTinterwordstretchfactor{4}
\providecommand\BIBentryALTinterwordspacing{\spaceskip=\fontdimen2\font plus
\BIBentryALTinterwordstretchfactor\fontdimen3\font minus
  \fontdimen4\font\relax}
\providecommand\BIBforeignlanguage[2]{{%
\expandafter\ifx\csname l@#1\endcsname\relax
\typeout{** WARNING: IEEEtran.bst: No hyphenation pattern has been}%
\typeout{** loaded for the language `#1'. Using the pattern for}%
\typeout{** the default language instead.}%
\else
\language=\csname l@#1\endcsname
\fi
#2}}

\bibitem{learning2019abbatematteo}
B.~Abbatematteo, S.~Tellex, and G.~Konidaris, ``{Learning to Generalize
  Kinematic Models to Novel Objects.}'' \emph{Proc.~of the Conference on Robot
  Learning (CoRL)}, 2019.

\bibitem{articulated-object-pose-estimation}
X.~Li, H.~Wang, L.~Yi, L.~J. Guibas, A.~L. Abbott, and S.~Song,
  ``Category-level articulated object pose estimation,'' \emph{Proc.~of the
  IEEE Conference on Computer Vision and Pattern Recognition (CVPR)}, 2020.

\bibitem{screwnet}
A.~Jain, R.~Lioutikov, and S.~Niekum, ``{ScrewNet: Category-Independent
  Articulation Model Estimation From Depth Images Using Screw Theory},''
  \emph{Proc.~of the IEEE Int.~Conf.~on Robotics \& Automation (ICRA)}, 2021.

\bibitem{shape2motion2019wang}
X.~Wang, B.~Zhou, Y.~Shi, X.~Chen, Q.~Zhao, and K.~Xu, ``{Shape2Motion: Joint
  Analysis of Motion Parts and Attributes From 3D Shapes},'' \emph{Proc.~of the
  IEEE Conference on Computer Vision and Pattern Recognition (CVPR)}, 2019.

\bibitem{articulated2021mittal}
M.~Mittal, D.~Hoeller, F.~Farshidian, M.~Hutter, and A.~Garg, ``{Articulated
  Object Interaction in Unknown Scenes with Whole-Body Mobile Manipulation},''
  \emph{arXiv}, 2021.

\bibitem{arduengo}
M.~Arduengo, C.~Torras, and L.~Sentis, ``{A Versatile Framework for Robust and
  Adaptive Door Operation with a Mobile Manipulator Robot},'' \emph{arXiv},
  2019.

\bibitem{where2act2021mo}
K.~Mo, L.~J. Guibas, M.~Mukadam, A.~Gupta, and S.~Tulsiani, ``{Where2Act: From
  Pixels to Actions for Articulated 3D Objects},'' in \emph{Proc.~of the IEEE
  Int.~Conf. on Computer Vision (ICCV)}, 2021.

\bibitem{vatmart2021wu}
R.~Wu, Z.~Guo, Q.~Fan, Y.~Zhao, X.~Chen, L.~Guibas, H.~Dong, Y.~Wang, T.~Wu,
  and K.~Mo, ``{VAT-Mart: Learning Visual Action Trajectory Proposals for
  Manipulating 3D ARTiculated Objects},'' \emph{Proc.~of the International
  Conference on Learning Representations (ICLR)}, 2021.

\bibitem{information_theoretic_mpc}
G.~Williams, N.~Wagener, B.~Goldfain, P.~Drews, J.~M. Rehg, B.~Boots, and E.~A.
  Theodorou, ``{Information theoretic MPC for model-based reinforcement
  learning},'' \emph{Proc.~of the IEEE Int.~Conf.~on Robotics \& Automation
  (ICRA)}, 2017.

\bibitem{giuseppe_mppi}
G.~M. Rizzi, J.~J. Chung, A.~R. Gawel, L.~Ott, M.~Tognon, and R.~Siegwart,
  ``Robust sampling-based control of mobile manipulators for interaction with
  articulated objects,'' \emph{IEEE Transactions on Robotics}, vol.~39, no.~3,
  2023.

\bibitem{object-size-estimation}
H.~Wang, S.~Sridhar, J.~Huang, J.~Valentin, S.~Song, and L.~J. Guibas,
  ``{Normalized Object Coordinate Space for Category-Level 6D Object Pose and
  Size Estimation},'' \emph{Proc.~of the IEEE Conference on Computer Vision and
  Pattern Recognition (CVPR)}, 2019.

\bibitem{kineverse2022}
A.~Rofer, G.~Bartels, W.~Burgard, A.~Valada, and M.~Beetz, ``Kineverse: A
  symbolic articulation model framework for model-agnostic mobile
  manipulation,'' \emph{IEEE Robotics and Automation Letters}, vol.~7, no.~2,
  pp. 3372--3379, 2022.

\bibitem{sturm2010}
J.~Sturm, A.~Jain, C.~Stachniss, C.~C. Kemp, and W.~Burgard, ``{Operating
  articulated objects based on experience},'' \emph{Proc.~of the IEEE/RSJ
  Int.~Conf.~on Intelligent Robots and Systems (IROS)}, 2010.

\bibitem{martinmartin2019}
R.~Martín-Martín and O.~Brock, ``{Coupled recursive estimation for online
  interactive perception of articulated objects},'' \emph{The International
  Journal of Robotics Research}, 2019.

\bibitem{jain2020}
A.~Jain and S.~Niekum, ``{Learning Hybrid Object Kinematics for Efficient
  Hierarchical Planning Under Uncertainty},'' \emph{Proc.~of the IEEE/RSJ
  Int.~Conf.~on Intelligent Robots and Systems (IROS)}, 2020.

\bibitem{mppi_control}
G.~Williams, A.~Aldrich, and E.~A. Theodorou, ``{Model Predictive Path Integral
  Control: From Theory to Parallel Computation},'' \emph{Journal of Guidance,
  Control, and Dynamics}, vol.~40, no.~2, pp. 344--357, 2017.

\bibitem{stein_var_mpc}
A.~Lambert, A.~Fishman, D.~Fox, B.~Boots, and F.~Ramos, ``{Stein Variational
  Model Predictive Control},'' \emph{Proc.~of the Conference on Robot Learning
  (CoRL)}, 2020.

\bibitem{ecological1979gibson}
J.~J. Gibson, \emph{The Ecological Approach to Visual Perception}.\hskip 1em
  plus 0.5em minus 0.4em\relax Houghton Mifflin, Boston MA, U.S.A, 1979.

\bibitem{affordances2020ardn}
P.~Ardon, E.~Pairet, K.~S. Lohan, S.~Ramamoorthy, and R.~P.~A. Petrick,
  ``{Affordances in Robotic Tasks -- A Survey},'' \emph{arXiv}, 2020.

\bibitem{learning2020nagarajan}
T.~Nagarajan and K.~Grauman, ``Learning affordance landscapes for interaction
  exploration in 3d environments,'' \emph{Proc.~of Advances in Neural
  Information Processing Systems (NeurIPS)}, 2020.

\bibitem{li2022ifrexplore}
Q.~Li, K.~Mo, Y.~Yang, H.~Zhao, and L.~Guibas, ``{IFR-Explore: Learning
  Inter-object Functional Relationships in 3D Indoor Scenes},'' in
  \emph{Proc.~of the International Conference on Learning Representations
  (ICLR)}, 2022.

\bibitem{learning2018zeng}
A.~Zeng, S.~Song, S.~Welker, J.~Lee, A.~Rodriguez, and T.~Funkhouser,
  ``{Learning Synergies Between Pushing and Grasping with Self-Supervised Deep
  Reinforcement Learning},'' \emph{Proc.~of the IEEE/RSJ Int.~Conf.~on
  Intelligent Robots and Systems (IROS)}, 2018.

\bibitem{pohl2020affordance}
C.~Pohl, K.~Hitzler, R.~Grimm, A.~Zea, U.~D. Hanebeck, and T.~Asfour,
  ``{Affordance-based grasping and manipulation in real world applications},''
  \emph{Proc.~of the IEEE/RSJ Int.~Conf.~on Intelligent Robots and Systems
  (IROS)}, 2020.

\bibitem{learning2020yenchen}
L.~Yen-Chen, A.~Zeng, S.~Song, P.~Isola, and T.-Y. Lin, ``{Learning to See
  before Learning to Act: Visual Pre-training for Manipulation},''
  \emph{Proc.~of the IEEE Int.~Conf.~on Robotics \& Automation (ICRA)}, 2020.

\bibitem{o2oafford2021}
K.~Mo, Y.~Qin, F.~Xiang, H.~Su, and L.~J. Guibas, ``{O2O-Afford:
  Annotation-Free Large-Scale Object-Object Affordance Learning},''
  \emph{Proc.~of the Conference on Robot Learning (CoRL)}, 2021.

\bibitem{interaction_tensor}
E.~Ruiz and W.~W. Mayol-Cuevas, ``{Geometric Affordance Perception: Leveraging
  Deep 3D Saliency With the Interaction Tensor.}'' \emph{Frontiers in
  Neurorobotics}, 2020.

\bibitem{3daffordancenet}
S.~Deng, X.~Xu, C.~Wu, K.~Chen, and K.~Jia, ``{3D AffordanceNet: A Benchmark
  for Visual Object Affordance Understanding},'' \emph{Proc.~of the IEEE
  Conference on Computer Vision and Pattern Recognition (CVPR)}, 2021.

\bibitem{Myers2015AffordanceDO}
A.~Myers, C.~L. Teo, C.~Ferm{\"u}ller, and Y.~Aloimonos, ``{Affordance
  detection of tool parts from geometric features},'' \emph{Proc.~of the IEEE
  Int.~Conf.~on Robotics \& Automation (ICRA)}, 2015.

\bibitem{ada_afford}
Y.~Wang, R.~Wu, K.~Mo, J.~Ke, Q.~Fan, L.~Guibas, and H.~Dong, ``{AdaAfford:
  Learning to Adapt Manipulation Affordance for 3D Articulated Objects via
  Few-shot Interactions},'' \emph{arXiv}, 2021.

\bibitem{sapien2020xiang}
F.~Xiang, Y.~Qin, K.~Mo, Y.~Xia, H.~Zhu, F.~Liu, M.~Liu, H.~Jiang, Y.~Yuan,
  H.~Wang, L.~Yi, A.~X. Chang, L.~J. Guibas, and H.~Su, ``{SAPIEN: A SimulAted
  Part-based Interactive ENvironment},'' \emph{Proc.~of the IEEE Conference on
  Computer Vision and Pattern Recognition (CVPR)}, 2020.

\bibitem{umpnet_xu_2022}
Z.~Xu, Z.~He, and S.~Song, ``{UMPNet: Universal manipulation policy network for
  articulated objects},'' \emph{{IEEE} Robotics and Automation Letters}, 2022.

\bibitem{qi2017pointnetplusplus}
C.~R. Qi, L.~Yi, H.~Su, and L.~J. Guibas, ``{PointNet++: Deep Hierarchical
  Feature Learning on Point Sets in a Metric Space},'' \emph{Proc.~of Advances
  in Neural Information Processing Systems (NeurIPS)}, 2017.

\bibitem{partnet2019mo}
K.~Mo, S.~Zhu, A.~X. Chang, L.~Yi, S.~Tripathi, L.~J. Guibas, and H.~Su,
  ``{PartNet: A Large-Scale Benchmark for Fine-Grained and Hierarchical
  Part-Level 3D Object Understanding},'' \emph{Proc.~of the IEEE Conference on
  Computer Vision and Pattern Recognition (CVPR)}, 2019.

\bibitem{contactdynamics2020}
A.~M. Castro, A.~Qu, N.~Kuppuswamy, A.~Alspach, and M.~Sherman, ``{A
  Transition-Aware Method for the Simulation of Compliant Contact with
  Regularized Friction},'' \emph{IEEE Robotics and Automation Letters}, 2020.

\bibitem{IKEA}
``{IKEA Furniture Models},'' \url{https://www.ikea.com/ch/en/planners},
  accessed: 2022-06-15.

\bibitem{asdf}
J.~Mu, W.~Qiu, A.~Kortylewski, A.~Yuille, N.~Vasconcelos, and X.~Wang,
  ``{A-SDF: Learning Disentangled Signed Distance Functions for Articulated
  Shape Representation},'' \emph{Proc.~of the IEEE Int.~Conf. on Computer
  Vision (ICCV)}, 2021.

\bibitem{weng2021captra}
Y.~Weng, H.~Wang, Q.~Zhou, Y.~Qin, Y.~Duan, Q.~Fan, B.~Chen, H.~Su, and L.~J.
  Guibas, ``{CAPTRA: CAtegory-level Pose Tracking for Rigid and Articulated
  Objects from Point Clouds},'' \emph{Proc.~of the IEEE Int.~Conf. on Computer
  Vision (ICCV)}, 2021.

\bibitem{pavlasek2020}
J.~Pavlasek, S.~Lewis, K.~Desingh, and O.~C. Jenkins, ``{Parts-Based
  Articulated Object Localization in Clutter Using Belief Propagation},''
  \emph{Proc.~of the IEEE/RSJ Int.~Conf.~on Intelligent Robots and Systems
  (IROS)}, 2020.

\bibitem{desingh2019}
K.~Desingh, S.~Lu, A.~Opipari, and O.~C. Jenkins, ``{Factored Pose Estimation
  of Articulated Objects using Efficient Nonparametric Belief Propagation},''
  \emph{Proc.~of the IEEE Int.~Conf.~on Robotics \& Automation (ICRA)}, 2019.

\bibitem{visualzeng2021}
V.~Zeng, T.~E. Lee, J.~Liang, and O.~Kroemer, ``{Visual Identification of
  Articulated Object Parts},'' \emph{Proc.~of the IEEE/RSJ Int.~Conf.~on
  Intelligent Robots and Systems (IROS)}, 2021.

\bibitem{ditto}
Z.~Jiang, C.-C. Hsu, and Y.~Zhu, ``{Ditto: Building Digital Twins of
  Articulated Objects from Interaction},'' \emph{Proc.~of the IEEE Conference
  on Computer Vision and Pattern Recognition (CVPR)}, 2022.

\bibitem{mask_r_cnn}
K.~He, G.~Gkioxari, P.~Dollar, and R.~Girshick, ``{Mask R-CNN},'' in
  \emph{Proceedings of the IEEE International Conference on Computer Vision
  (ICCV)}, 2017.

\bibitem{icp}
``{Open3D ICP Registration},''
  \url{http://www.open3d.org/docs/release/tutorial/pipelines/icp_registration.html},
  accessed: 2022-06-22.

\end{thebibliography}

\clearpage

\section*{APPENDIX}

\subsection{Controller}
\label{sec:mppi}
\paragraph{Sampling-based controller}
In this section, we state the equations of our sampling-based controller based on \cite{giuseppe_mppi}. Let $\configRobot \in \nR{n_q}$ and $\dconfigRobot \in \nR{n_q}$ be the configuration of the agent $\robot$ and its time derivative, where $n_q \in \nN{}_{>0}$ is the agent's degrees of freedom (DOF) \---- in our implementation $n_q=10$. Similarly, let $\configObject \in \nR{n_o}$ and $\dconfigObject \in \nR{n_o}$ be the configuration of the object and its time derivative, where $n_o \in \nN{}_{>0}$ is the object's DOF \---- in our implementation $n_o=1$. The controller state vector is defined as:
\begin{equation}
    \state = [\configRobot^T, \dconfigRobot^T, \configObject^T, \dconfigObject^T]^T \in \nR{2(n_q+n_o)}.
\end{equation}

The controller defines several stage cost function components. The first component is an object cost, which encodes the object manipulation objective:
\begin{equation}
    c_{\configObject}(\configObject, \configObject^*; \matr{W}_{\configObject}) = ||\configObject - \configObject^*||^2_{\matr{W}_{\configObject}},
\end{equation}
where $\configObject^*$ is the target object configuration and $\matr{W}_{\configObject} \in \nR{n_o \times n_o}$ is a diagonal weight matrix. The second component is the pose reach cost, which encodes the distance of the end-effector to the reference interaction pose $\pose^* = \text{(position }\point^*\text{, orientation }\orientation^*)$:
\begin{equation}
    c_t(\configRobot, \pose^*; \matr{W}_t) = ||\log(\pose(\configRobot) - \pose^*)||^2_{\matr{W}_t},
\end{equation}
where the actual end-effector pose $\pose(\configRobot)$ is computed via forward kinematics from the joint configuration vector $\configRobot$ and $\matr{W}_t \in \nR{6 \times 6}$ is a diagonal weight matrix. \\

A number of surrogate objectives are defined to encode agent limits and constraints:
\begin{itemize}
    \item Collision cost:
    \begin{equation*}
        c_c(\configObject, \configRobot; w_c) = 
        \begin{cases}
            w_c & \text{if agent-object collision} \\
            0 &   \text{otherwise}
        \end{cases}
    \end{equation*}
    where $w_c \in \nR{}$.
    \item Joint limit cost
    \begin{align*}
    c_j(\configRobot; w_j, \matr{W}_j) = &\mathbbm{1}[\configRobot > \configRobot _{\text{upper}}](w_j + ||\configRobot  - \configRobot _{\text{upper}}||^2_{\matr{W}_j}) + \\ &\mathbbm{1}[\configRobot  < \configRobot _{\text{lower}}](w_j + ||\configRobot  - \configRobot _{\text{lower}}||^2_{\matr{W}_j})
    \end{align*}
    where $\configRobot_{\text{upper}}$ and $\configRobot_{\text{lower}}$ are the upper and lower joint limits, $w_j \in \nR{}$ and $\matr{W}_j \in \nR{n_q \times n_q}$.
    \item Arm reach cost
    \begin{equation*}
    c_a(\configRobot; w_a, w_{as}) = \mathbbm{1}[r(\configRobot) > r_{\text{max}}](w_a + w_{as}(r(\configRobot) - r_{\text{max}}))
    \end{equation*}
    where the current reach $r$ is calculated from forward kinematics from the joint state $\configRobot$ and $w_a, w_{as} \in \nR{}$ are weights.
\end{itemize} 

The controller defines two modes of operation $\mppiMode \in \{1,2\}$, corresponding to two stages in object manipulation \emph{reaching} and \emph{interacting}. In the first phase, the end-effector reaches the target interaction pose while avoiding contact with the object. This is achieved by setting the stage cost as:
\begin{equation}
    l(\state_i, \cinput_i; \mppiMode=1) = c_c + c_j + c_a + c_t.
\end{equation}
In the second phase, the agent interacts with the object and moves it towards the target state:
\begin{equation}
    l(\state_i, \cinput_i; \mppiMode=2) = c_{\configObject} + c_j + c_a + c_t.
\end{equation}

\paragraph{Low-level controller}
We use a dynamically compensated low-level proportional-integral controller to convert the velocity references $\cinput$ from the sampling-based controller to joint torque commands $\jointtorque$. Let $\configRobotDesired$ be the desired agent configuration, calculated by integrating the velocity references over the time interval $d\mtime$:
\begin{equation*}
    \configRobotDesired = \configRobot + \cinput d\mtime.
\end{equation*}
Let $\massMatrix \in \nR{n_q \times n_q}$ be the matrix of inertia of the agent. The Coriolis and gravity terms of the system dynamics are denoted as $\coriolis$ and $\vect{g}(\configRobot)$, respectively. The joint torque vector is calculated as:
\begin{equation}
    \jointtorque = \massMatrix \ddconfigRobotDesired + \coriolis \dconfigRobotDesired + \vect{g}(\configRobot) - \matr{K}_D \dconfigRobotError - \matr{K}_I \int_{0}^{t} \dconfigRobotError d\tau,
\end{equation}
with the auxiliary error variable $\configRobotError = \configRobot - \configRobotDesired$ and $\matr{K}_D$ and $\matr{K}_I$ being positive diagonal gain matrices.

\paragraph{Task scheduler}
\label{sec:task_scheduler}
The task scheduler selects the appropriate mode of the sampling-based controller, which can either be \emph{reaching} ($\mppiMode=1$) or \emph{interacting} ($\mppiMode=2$). It also regulates updates in the interaction pose $\pose$. Given the current state observation $\stateobs$, the target object configuration $\objTarget$ and the current pose reference $\pose$, the task scheduler implements the following simple heuristic:
\begin{itemize}
    \item At the beginning of the interaction, update the reference pose and go into \emph{reaching} mode ($m=1)$. 
    \item During the interaction:
    \begin{itemize}
        \item If in \emph{reaching} mode and the current reference pose was reached by the end-effector, change into \emph{interacting} mode ($\mppiMode=2$).
        \item If in \emph{interacting} mode and the object configuration did not improve for a set number of seconds (5s in simulation, 1s in real-life testing), lift the robot arm to prevent camera occlusion while recording a new point cloud, trigger an update of the reference pose and return to \emph{reaching} mode.
        \item If $|\configObject - \objTarget| \leq 5^\circ$, stop the interaction as the task is fulfilled.
    \end{itemize}
\end{itemize}
The task scheduling described above allows the agent to change the interaction pose when the current one is not advantageous anymore as well as to try again if the current interaction pose failed to lead to any improvement in the object configuration. As both the sampling-based controller and the pose module are non-deterministic and the agent moves in between the pose updates, the same initial conditions of task and object might result in different behavior, i.e. success where before the agent failed.

\subsection{Affordance Definitions}
\label{sec:affordance_definition}
This section gives an overview of the equations used for the pose module. Let $\reward(\state_{0:N}, \cinput_{0:N-1}; \objTarget)$ be a reward function that, given the the target object configuration $\objTarget$, maps a trajectory of system states $\{\state_{0:N}\}$ and controller inputs $\{\cinput_{0:N-1}\}$ to a binary reward value:
\begin{equation}
    \reward(\state_{0:N}, \cinput_{0:N-1}; \objTarget)=
    \begin{cases}
    1 & \text{if} \quad |\configObject_0 - \objTarget| - |\configObject_N - \objTarget| > \theta \\
    0 & \text{otherwise},
    \end{cases}
    \end{equation}
where $\theta \in \nR{}_{>0}$ is a success threshold (in our implementation $\theta = 5^\circ$).

Given an object, a target configuration $\objTarget$ and an initial state $\state_0$, we  define the affordance $\aff_{\reward}$ of an interaction pose ($\point, \orientation$) as the expected value of $\reward$ over all the state-input trajectories induced by the controller $\controller(\state; \objTarget, \point, \orientation)$ on the agent-object system from initial state $\state_0$. In our method, the object and the initial state are implicitly represented by the point cloud which is passed through the PointNet++ encoder \cite{qi2017pointnetplusplus} to generate point-wise feature vectors $\pfeature(\point, \state_0, \text{object})$. Therefore we use the following definition of affordance $\aff_{\reward}$:

\begin{equation}
\aff_{\reward}(\point, \orientation, \pfeature, \objTarget) :=  \mathop{\mathbb{E}}_{\state_i, \controller(\state_i)}\left( \reward \left( \state_{0:N}, \controller(\state_{0:N-1}); \objTarget \right) \right).
\end{equation}

Following this affordance definition, the optimal interaction pose ($\point^*, \orientation^*$) is given by maximizing the affordance function $\aff_\reward$:
\begin{equation}
\point^*, \orientation^* = \arg\max_{\point, \orientation} \aff_\reward(\point, \orientation, \pfeature, \objTarget).
\end{equation}

Solving this equation for all possible interaction poses is computationally intractable, therefore we define auxiliary functions to optimize the position and orientation in a hierarchical manner. The orientation proposal distribution $\orprop_\reward$ generates orientation samples $\orientation \approx \orientation^*$ that approximate the optimum for a given interaction point $\point$:
    \[\orientation \sim \orprop_\reward(\point, \pfeature, \objTarget).\]
    
The actionability function $\act_{\reward}$ is defined as the point-wise expected value of the affordance over interaction orientations sampled from an orientation proposal distribution $\orprop_\reward$: 
\begin{equation}
    \label{actionability_definition}
    \act_{\reward}(\point, \pfeature, \objTarget) = \mathop{\mathbb{E}}_{\orientation \sim \orprop} \aff_{\reward}(\point, \orientation, \pfeature, \objTarget).
\end{equation}
This allows us to first obtain the interaction point $\hat{\point}$ by sampling the actionability:
        \begin{equation}
        \hat{\point} = \arg\max_{\point} \act_{\reward}(\point, \pfeature, \objTarget).
        \end{equation}
        
Next, a set of orientation proposals are sampled from $\orprop_\reward$.
Finally the affordance function $\aff_\reward$ is used to score the orientation proposals and sample the highest-scoring one:
\begin{equation}
\hat{\orientation} = \arg\max_{\orientation \sim \orprop} \aff_{\reward}(\hat{\point}, \orientation, \vect{f_{\hat{p}}}, \objTarget).
\label{equation:p_and_q_star}
\end{equation}

This hierarchical procedure does not necessarily output the globally optimal interaction pose. Instead, the quality of the solution heavily depends on the orientation sampling function $\orprop_\reward$. If $\orientation \sim \orprop$ only and always samples the optimal interaction orientation, the procedure will yield the global pose optimum.

\subsection{Simulation Settings and Training}
\paragraph{Simulation settings}
Dynamics are simulated using a physics simulation based on the commercial software RAISIM. We set the simulation timestep to $0.0015$ s. The range of motion of the object articulation joint is set to $90^\circ$. To increase simulation speed, we do not check for collisions between the two object links. We set a damping coefficient of $20$ and friction of $40$ at the object articulation joint. Between surfaces of contact bodies, we set a low friction coefficient ($0.01$).

\paragraph{Training data collection}
\label{app:sec:training_data_collection}
Each object in the PartNet Mobility dataset is scale-normalized within a unit sphere. For each simulation instance, we sample an object from the category \emph{oven} or \emph{dishwasher} and apply a scale factor $\in[45\%, 55\%]$ of the normalized scale. This ensures a realistic relative scale between our object and agent models, and serves as augmentation. The position of the object, measured from the world origin to the center of the object, is sampled uniformly between the following bounds ($x=0 \text{ m}, y\in[-0.3 \text{ m}, 0.3\text{ m}], z\in[0.75 \text{ m}, 0.95 \text{ m}]$). We initialize the object articulation joint at rest $50\%$ of the time (i.e. $0^\circ$ for the \textit{open} task, $90^\circ$ for the \textit{close} task) and in an intermediate state for the other 50\% of initializations. We sample the initial and target configuration within the articulation bounds and according to the task, with a minimum task distance $\Delta \configObject = | \objTarget - \configObject_0 |$ of $20^\circ$.

\begin{figure}[ht]
    \centering
    \includegraphics[width = 0.47\textwidth]{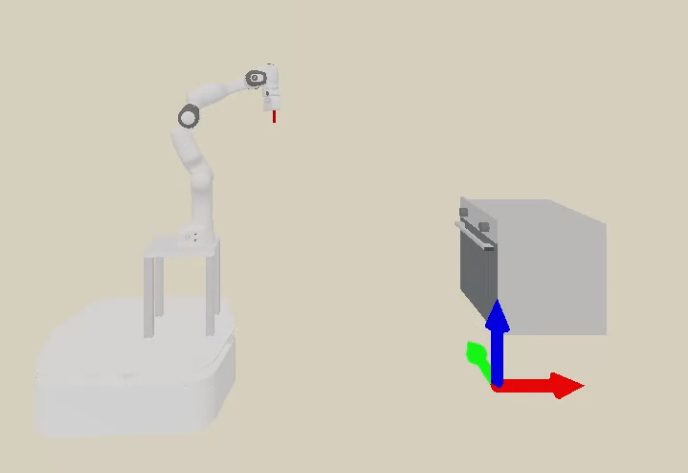}
    \caption{Rendered simulation environment with world axes (x: red, y: green, z: blue) for reference.}
    \label{fig:simulation_env_example}
\end{figure}

The agent is initialized facing the world origin, with a uniformly sampled distance $\in[1.75\text{ m}, 3.0 \text{ m}]$ and base rotation angle $\arctan(y, x)\in[-20^\circ, 20^\circ]$. During the simulated interaction, the agent is given $30$s to move to the pose reference (i.e. mode $\mppiMode = 1$ of the sampling-based controller) and $5$s to interact with the object (i.e. $\mppiMode = 2$). A simulation is considered successful if the object configuration was moved by at least $5^\circ$ towards the task. Figure \ref{fig:simulation_env_example} shows our simulation environment, complete with axes for reference. We collect $12000$ simulations for each task, which takes around $15$ hours per task on a commercial CPU (Intel Core i7-7700HQ quad-core processor).

\paragraph{Network training}
\label{app:sec:network_params}
We use an Adam optimizer with an initial learning rate of $1\mathrm{e}{-4}$ and batch size of $10$. To ensure convergence, we employ both learning rate scheduling (patience of $5$ epochs, discount factor $0.2$, minimum improvement of $2\%$, minimum learning rate of $1\mathrm{e}{-6}$) and early stopping (patience of $12$ epochs, minimum improvement of $5\%$). Fully training a network usually requires between $30$ and $50$ epochs, and takes around $3$ hours on a low-grade GPU (NVIDIA GTX $1050$ Ti, with $4$ GB of dedicated RAM).

\paragraph{Evaluation setup}
\label{app:sec:pose_testing_params}
For evaluation, the simulation is initialized in the same manner as during training data collection. In the first experiment of the paper, the agent is given $40$s to move to the pose reference and $30$s to interact with the object. The sample success rate is evaluated, which is the percentage of successful pose references and success is defined as a movement of the object configuration $\configObject$ by at least $5^\circ$ towards the target $\objTarget$:
\begin{equation}
\text{sample success rate} = \frac{\text{n successful proposals}}{\text{n proposals}}.
\end{equation}
We also report the sample reach rate, which is the percentage of pose references that are reachable by the agent if given as a target pose to the sampling-based controller:
\begin{equation}
\text{sample reach rate} = \frac{\text{n reached proposals}}{\text{n proposals}}.
\end{equation}

In the second experiment, which evaluates the effect of the closed-loop dynamic pose update, the agent is given $40$s to move to the pose reference and $300$s to interact with the object. We report the task success rate, where a task is considered successful if the final object configuration $\configObject$ is within $5^\circ$ of the target $\objTarget$: 
\begin{equation}
\text{task success rate} = \frac{\text{n successful tasks}}{\text{n tasks}}.
\end{equation}

In the third experiment, the pipeline is tested on objects of the category \emph{safe}, \emph{washing machine} and \emph{table}, applying a scale factor of 50\%, 50\% and 100\% respectively to ensure a realistic proportion between agent and object. Following VAT-Mart \cite{vatmart2021wu}, we first uniformly sample an object category, and then sample a shape from this category. We sample a task $\Delta \configObject = | \objTarget - \configObject_0 |$ in a range of $[10^\circ, 70^\circ]$ for revolute joints and $[0.1, 0.7]$ for prismatic joints. The initial object configuration $\configObject_0$ is randomly sampled such that it lies within the articulation joint bounds given the task. We report the task success rate, where we define a successful articulation joint value equivalently to VAT-Mart within a tolerance of 15\% of the task $\Delta \configObject = | \objTarget - \configObject_0|$.

\subsection{Network Sensitivity Study}
\label{sec:repeatability}
\begin{figure} [ht]
   \centering
   \includegraphics[width=0.5\textwidth]{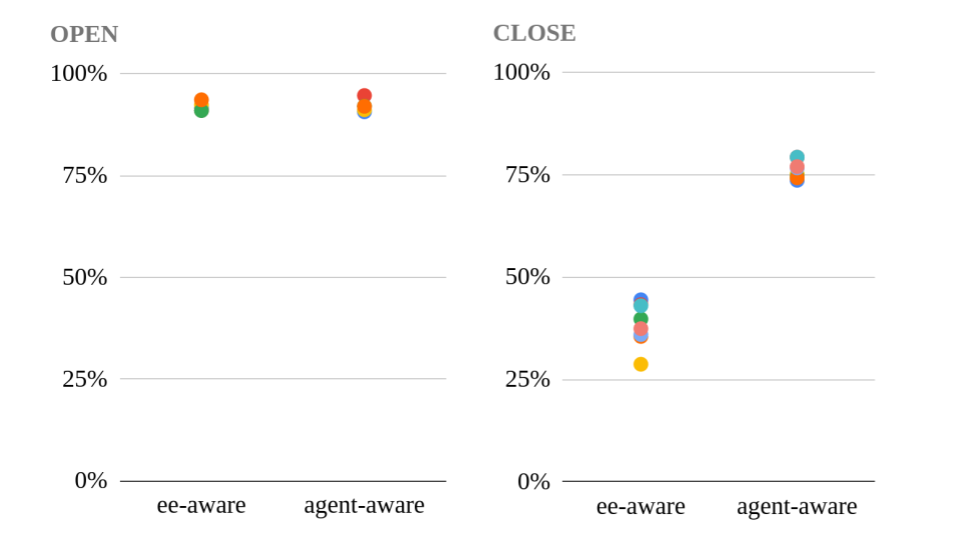}
  \caption{Sample success rate of the end-effector-aware and the agent-aware networks for the \emph{open} task across $5$ training runs (left) and for the \emph{close} task (right) across $8$ training runs with different random seeds (indicated by the colors).}
  \label{chart:ee_vs_agent}
\end{figure}

We identify three sources of randomness in the results of our interaction trials. Firstly, the random seed used for initialization of the artificial neural networks causes variability in the output of the network. Secondly, in our testing protocol the initial state and target object configuration are sampled at random. The third source of randomness lies in the stochastic nature of the sampling-based controller. To analyse these effects, we use the training data from the first experiment to train multiple networks initialized with different random seeds. The results are evaluated on $500$ interaction trials for each task and model (end-effector-aware and agent-aware). Figure \ref{chart:ee_vs_agent} reports the sample success rate of the two models for the \emph{open} task across $5$ training runs and for the \emph{close} across $8$ training runs. We observe that for the \emph{open} task, the performance of the networks is repeatable. In the \emph{close} task on the other hand, the end-effector-aware networks are far more inconsistent. This could be caused by the fact that behavior learned during training with a disembodied gripper doesn't transfer consistently to the testing task with the full agent. Overall, this analysis shows that the performance of the agent-aware pipeline is consistent over multiple training runs, and that therefore the influence of the stochasticity in controller and neural network is minimal.

\begin{figure} [ht]
   \centering
   \includegraphics[width=0.48\textwidth]{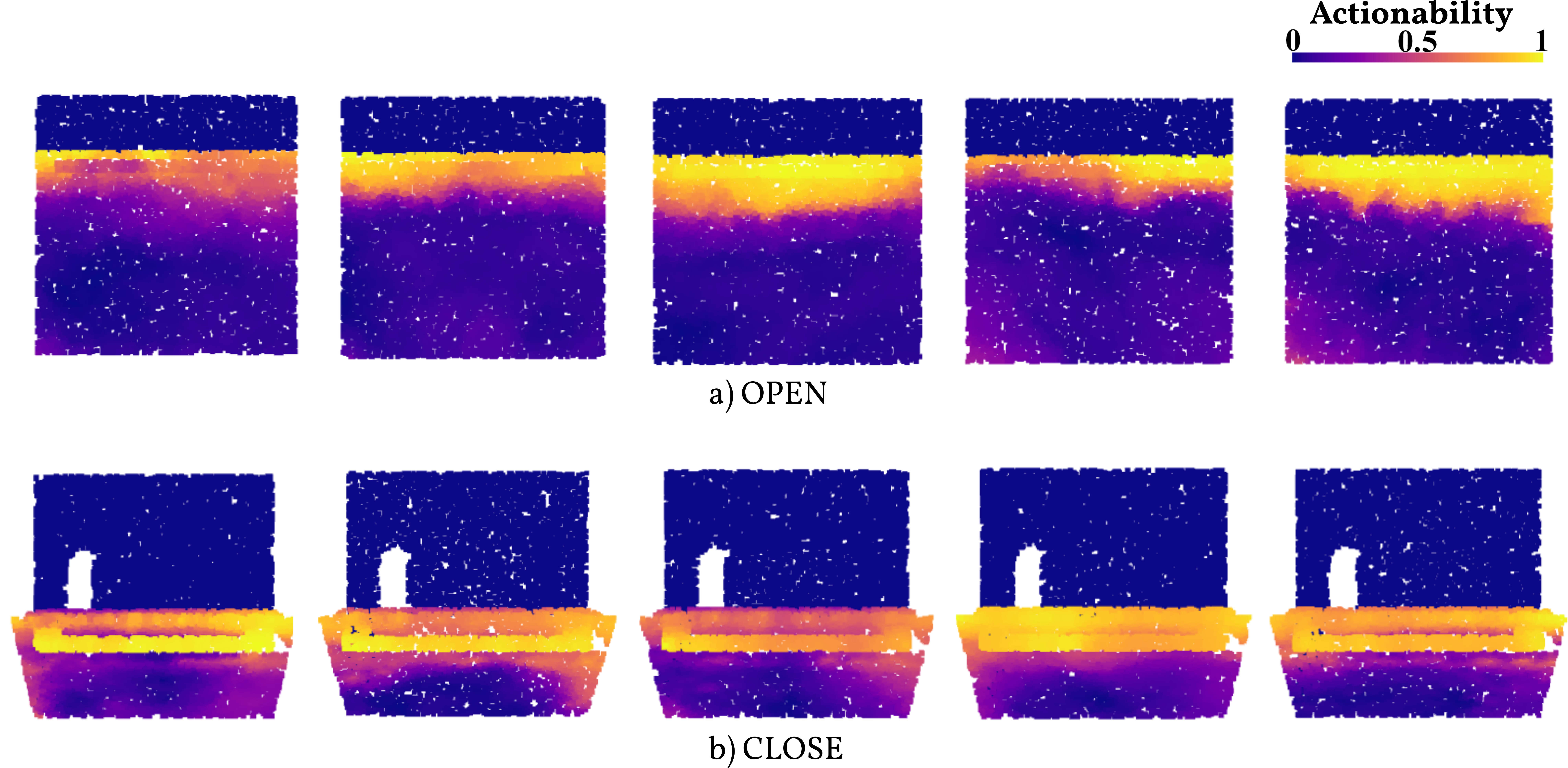}
  \caption{The predicted actionability scores for networks trained with different random seeds on (a) the \emph{open} and (b) the \emph{close} task.} 
  \label{image:different_networks}
\end{figure}

We analyzed the predictions of networks trained with different random seeds and observed that while the test-time performance is consistent across different networks, the spread of the predicted actionabilities varies (c.f. Fig. \ref{image:different_networks}). This suggests that the networks converge to different local minima, which could be caused by the fact that our training data is relatively sparse and has a stochastic component due to the controller. Additionally, it should be noted that at test time only the points with the highest actionability are sampled, therefore the distribution of the intermediate-level actionability scores does not have a direct influence on the executed interaction.  

\subsection{Hardware experiments}
\label{sec:hardware}

\begin{figure*}[ht]
\centering
\includegraphics[width = 0.85\textwidth]{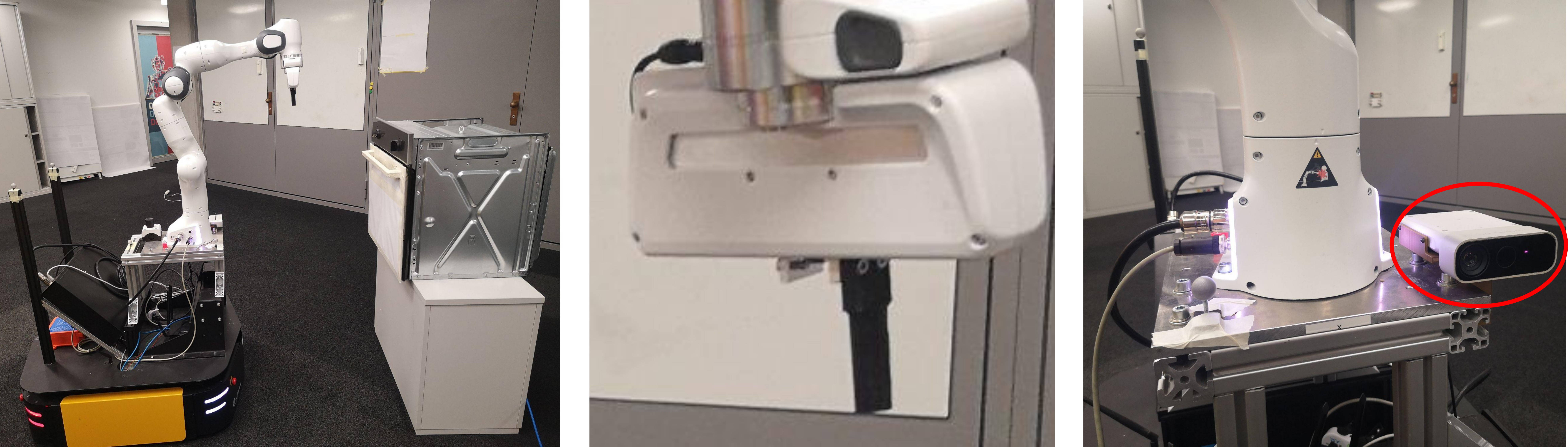}
\caption{Real-world setup showing the robotic platform and target object (left), the end-effector consisting of a single fixed finger (center) and the placement of the RGB-D Camera (right).}
\label{img:realworld_setup}
\end{figure*}

The hardware experiments were conducted with an omnidirectional platform (Clearpath Ridgeback) with a 7 degree-of-freedom robotic arm (Franka Emika Panda) mounted on top. The selected target object is an oven which was placed on a workbench, as seen in Fig. \ref{img:realworld_setup}. As it is required by our control pipeline, we build an articulation and collision model of the oven from real-world measurements.

The sampling-based controller is run on an onboard computer (Intel Core i7-8550U quad-core processor) at a frequency of $66$Hz, and a proportional-integral velocity controller computes joint torque commands for the arm at $1000$Hz. The Ridgeback base is controlled directly in velocity-space at a frequency of $50$Hz. The task scheduler is run at $30$Hz. The pose module is only run when a pose update is triggered, where each pose update requires roughly $1$s of computational time. The pose module and task scheduler are run on a separate laptop (Intel Core i7-7700HQ quad-core processor). The two devices communicate wirelessly over the network using the ROS framework.

The sampling-based controller requires an accurate state estimation. The pose of the Ridgeback mobile base is tracked using an OptiTrack motion capture system, while for base velocity we use odometry data. At startup, we calibrate the oven position in the global reference frame. To obtain the articulation joint angle, filtered linear acceleration and angular velocity are continuously extracted from an inertial measurement unit (IMU) mounted on the back of the oven door. The joint velocity is estimated by projecting the angular velocity measurements onto the articulation rotation axis, while joint position is estimated by integrating velocity.

\begin{figure*}[ht]
   \centering
   \includegraphics[width=0.9\textwidth]{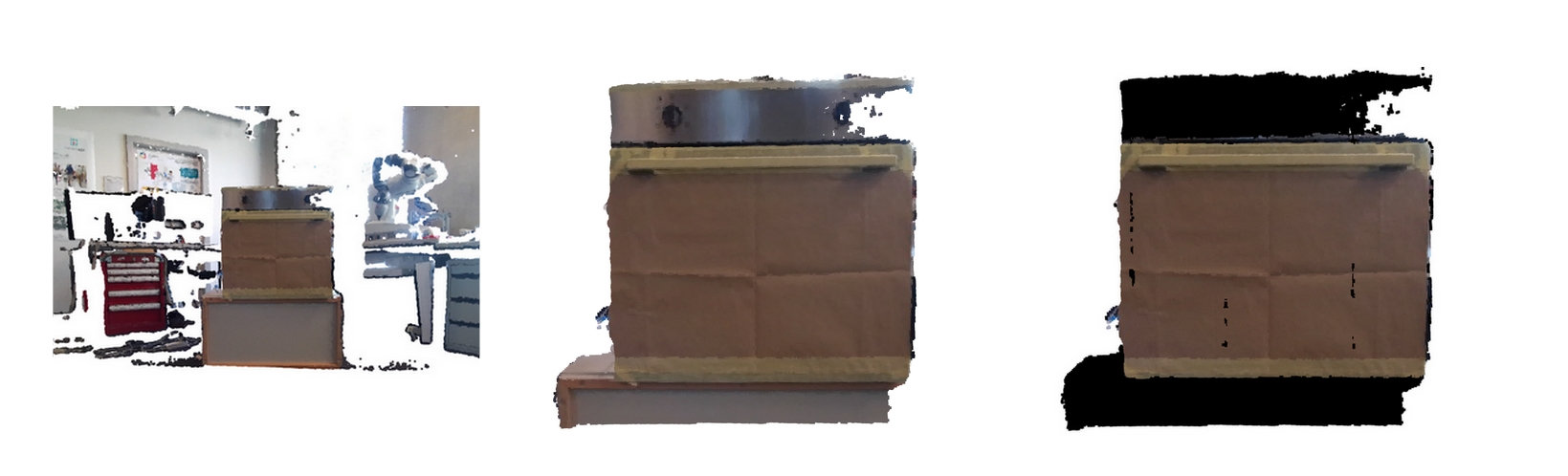}
   \caption{Segmentation process: scene point cloud (left), object point cloud (center) and part segmentation mask (right).}
   \label{image:segmentation_process}
\end{figure*}

Point cloud data is collected by an Azure Kinect sensor mounted on the robot base (Figure~\ref{img:realworld_setup}). While state-of-the-art object segmentation, e.g. a neural network such as Mask R-CNN \cite{mask_r_cnn}, could be implemented for segmentation, we used a simple heuristic which exploits the knowledge about the position and size of the object. The segmentation of the object from the scene is implemented using bounding boxes. For part segmentation, the Iterative Closest Point (ICP) algorithm is used to closely match the real-world point cloud to an expected point cloud obtained from our simulator/rendering setup \cite{icp}. Once the two point clouds are matched, the simulated point cloud is converted to a rough voxel map with a voxel size of $50$mm and the real-world point cloud is segmented by checking if its points are contained in the voxel map. Example images from this process can be found in Figure~\ref{image:segmentation_process}.

\begin{figure*} [ht]
   \centering
   \includegraphics[width=0.9\textwidth]{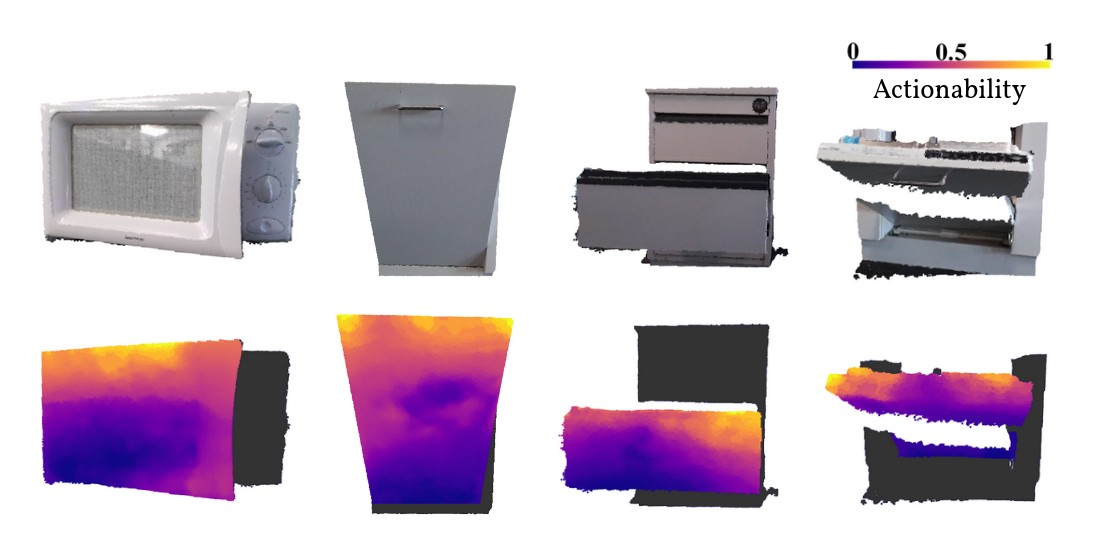}
   \caption{Real-world point-clouds of unseen articulated objects (top row) and the respective predicted actionability scores (bottom row).
    }
    \label{image:realworld_objects}
\end{figure*}

While the real-world experiment shows that our proposed system performs well on the oven, we would also like to showcase its generalization capabilities. Fig. \ref{image:realworld_objects} depicts the point cloud of multiple articulated objects with the corresponding affordance prediction. It shows that despite the partial and noisy point cloud of unseen objects, the affordance prediction module creates reasonable interaction scores.

\end{document}